# ENTERPRISE LARGE LANGUAGE MODEL EVALUATION BENCHMARK

Liya Wang[1], David Yi[1], Damien Jose[1], John Passarelli[1], James Gao[1], Jordan Leventis[1], and Kang Li[1]

[1]Atlassian
lwang10@atlassian.com

## ABSTRACT

*Large Language Models (LLMs) have demonstrated promise in boosting productivity across AI-powered tools, yet existing benchmarks like Massive Multitask Language Understanding (MMLU) inadequately assess enterprise-specific task complexities. We propose a 14-task framework grounded in Bloom's Taxonomy to holistically evaluate LLM capabilities in enterprise contexts. To address challenges of noisy data and costly annotation, we develop a scalable pipeline combining LLM-as-a-Labeler, LLM-as-a-Judge, and corrective retrieval-augmented generation (CRAG), curating a robust 9,700-sample benchmark. Evaluation of six leading models shows open-source contenders like DeepSeek R1 rival proprietary models in reasoning tasks but lag in judgment-based scenarios, likely due to overthinking. Our benchmark reveals critical enterprise performance gaps and offers actionable insights for model optimization. This work provides enterprises a blueprint for tailored evaluations and advances practical LLM deployment.*

## 1. INTRODUCTION

Large Language Models (LLMs) are transforming enterprise operations by automating tasks such as data analysis, code generation, and document creation. These capabilities not only increase efficiency but also inform strategic resource allocation. Moreover, recent advances in open-source LLMs have brought their performance on par with proprietary models, offering cost-effective solutions that strengthen enterprise AI ecosystems.

Robust benchmark datasets are essential for reliably evaluating the performance of fine-tuned LLMs. Such datasets should be diverse, novel, and sufficiently challenging [1], enabling engineers to determine the most suitable base models and identify areas where performance gaps still exist.

While multiple benchmarks such as Multitask Language Understanding (MMLU) [2], Big-Bench [3], ARC [4], Holistic Evaluation of Language Models (HELM) [5], MT-Bench and Chatbot Arena [6] exist, many of these often fail to account for the unique characteristics of enterprise applications. Additionally, these benchmarks are becoming saturated, suggesting that current evaluation methods may no longer effectively distinguish between newer, more advanced models and their predecessors.

Efforts to establish benchmarks for specific domains are evident across various fields. In the medical field, benchmarks including MedQA [7], MedQA-CS [8], MedMCQA [9], WorldMedQA-V [10], PubMedQA [11] and CMB [12] have been introduced. The financial sector has seen the introduction of benchmarks such as FinBen [13], FinEval [14], FLUE [15], BBT-Fin [16], XuanYuan 2.0 [17] and PIXIU [18]. In the legal realm, various benchmarks have been developed, such as LexGlue [19], LBOX OPEN [20], legal-bench [21], and LawBench [22]. In the realm of Chinese language evaluation, benchmarks like CMMLU [23], GAOKAO [24], and C-Eval [25] have been developed. For other languages, initiatives such as PersianMMLU [26], M3Exam [27] and AGIEval [28] have been proposed.

Our research aims to create a comprehensive enterprise evaluation benchmark dataset to assess the strengths and limitations of current LLMs in enterprise context. This will provide insights for LLM post-training and serve as a cost-saving blueprint for similar organizations.

## 2. Related Work

### 2.1. LLM-as-a-Labeler

In many enterprises, raw, unlabeled, and unstructured text data is abundant, while labeled data remains scarce. Traditionally, data annotation has relied on human labelers, a process that is time-consuming, costly, subjective, and difficult to scale. With the advent of LLMs like GPT-4o, organizations have a promising opportunity to revolutionize data annotation [29], [30], [31], [32].

LLMs offer several advantages for data annotation. They significantly enhance efficiency by rapidly processing large volumes of data, thereby reducing the time required to produce labelled datasets. They also improve consistency and objectivity, providing standardized annotations that minimize bias. Moreover, LLMs are highly scalable, enabling them to handle increasing data volumes without a corresponding increase in time or cost. They can flexibly generate data to meet specific requirements, such as identifying disease names in the medical field. Additionally, LLMs can adapt and learn from new data, enhancing annotation accuracy over time.

When combined with carefully crafted prompts or retrieval-augmented generation (RAG) techniques, LLMs can understand context and semantics with remarkable precision, supporting various annotation tasks. For example, they have been employed to generate instructions [33]; responses [17]; question-answering pair [34]; reasoning data [35], [36]; pairwise preference data [37], [38], [39], [40]; and textual feedback [41]. They are also effective in multiple-choice question answering (MCQA) [42] and in assigning confidence scores to assess annotation reliability [43].

In summary, incorporating LLMs into data annotation processes reduces costs, enhances data quality, and empowers enterprises to leverage their data more effectively.

### 2.2. Retrieval Augmented Generation (RAG)

LLMs are typically trained on historical data and often lack access to proprietary knowledge unique to individual enterprises. This limitation can lead to inaccuracies or hallucinations in their outputs. Retrieval-Augmented Generation (RAG) [44] addresses these limitations by incorporating domain-specific data, particularly private data, to tailor LLM applications to meet specific enterprise needs. LLMs enhanced with RAG offer several key benefits, including increased professionalism and timeliness, better alignment with domain experts, reduced likelihood of hallucinations, and improved controllability and explainability [45].

Recent advances have led to the development of several cutting-edge RAG techniques designed to overcome the limitations of basic RAG. For instance, MemoRAG [46] enhances retrieval by incorporating long-term memory capabilities, while Adaptive-RAG [47] dynamically selects strategies based on query complexity. Self-RAG [48] selectively retrieves knowledge and employs a critic model to determine whether retrieval is necessary. Similarly, Self-Knowledge guided Retrieval (SKR) [49] leverages both internal and external knowledge by extracting self-knowledge from LLMs. Retrieval-Augmented Language Models (RALMs) ) [50] integrate a natural language inference (NLI) model to identify and filter out irrelevant context, thereby enhancing robustness. Another innovative approach is SAIL [51], which enables fine-tuned language models to source, denoise, and reason using a combination of relevant and distracting search results. Additionally, Corrective Retrieval-Augmented Generation (CRAG) [52] integrates self-assessment, evaluation mechanisms, and large-scale web searches for document retrieval, collectively improving output quality.

### 2.3. LLM-as-a-Judge

Evaluating the outputs of LLMs poses significant challenges due to their complexity, subjective nature, and the wide variety of tasks they perform. Two common evaluation methods are used: automatic and human evaluation. Automatic evaluation, often referred to as "LLM-as-a-Judge," is faster, more cost-effective, and potentially more reliable than human evaluation (e.g., see [53], [54], [55], [6], [56], [57], [58] etc.). Currently, the method is commonly employed for a wide range of tasks including scoring, preference ranking, and candidate selection.

Despite its widespread use, there remains a disparity between LLM-as-a-Judge and human evaluation. Several techniques have been proposed to enhance the effectiveness of LLM-as-a-Judge. For example, PandaLM [59] enables reproducible and automated language model assessment by training an LLM to act as the "judge" in evaluating different models. Some methods incorporate pre-defined rules into prompts, directing the LLMs to adhere to explicit guidelines [60]. Additionally, swapping the order of two responses can help mitigate positional bias [61]. REVISEVAL leverages the text revision capabilities of LLMs to adaptively revise responses, treating the revised text as the reference (response-adapted reference) for subsequent evaluation [62]. G-Eval [63] adopts the Chain-of-Thoughts (CoT) approach [64] for better evaluation which has been widely adopted for its flexible creation of task-specific metrics.

LLM-as-a-Judge also faces certain challenges. For example, it has been observed [6], [58] that even with CoT prompts, LLM-as-a-Judge are sensitive to prompt complexity and length, and a tendency toward leniency. To address this, a reference-guided method has been proposed, where the LLM-as-a-Judge first generates a response independently based on given instructions and then uses this response as a reference in the evaluation prompt. Further research is needed to address those limitations.

## 3. BENCHMARKING METHODOLOGY

In this section, we offer an in-depth explanation of the principles guiding the design of our enterprise benchmark and the selection of test tasks. When compiling our enterprise benchmark dataset, we adhere to the traditional evaluation framework outlined in Figure 1 [65]. This framework composes of three parts: 1) what to evaluate: task construction; 2) where to evaluate: data construction; 3) how to evaluate: evaluation metrics calculation. Next, we present more details on implementing these three parts.

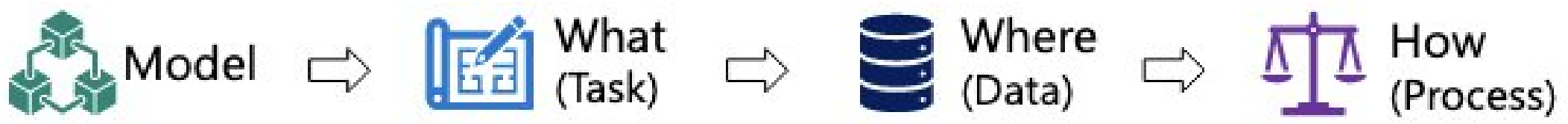

Figure 1. Evaluation pipeline [65]

### 3.1. Evaluation Tasks

To evaluate LLMs effectively, it is crucial to assemble a diverse set of tasks. Our extensive understanding of enterprise use cases has allowed us to categorize them into 14 distinct tasks, which we believe are commonly encountered across our operations. To systematically organize these tasks, we have chosen Bloom's Taxonomy [66] as an optimal framework.

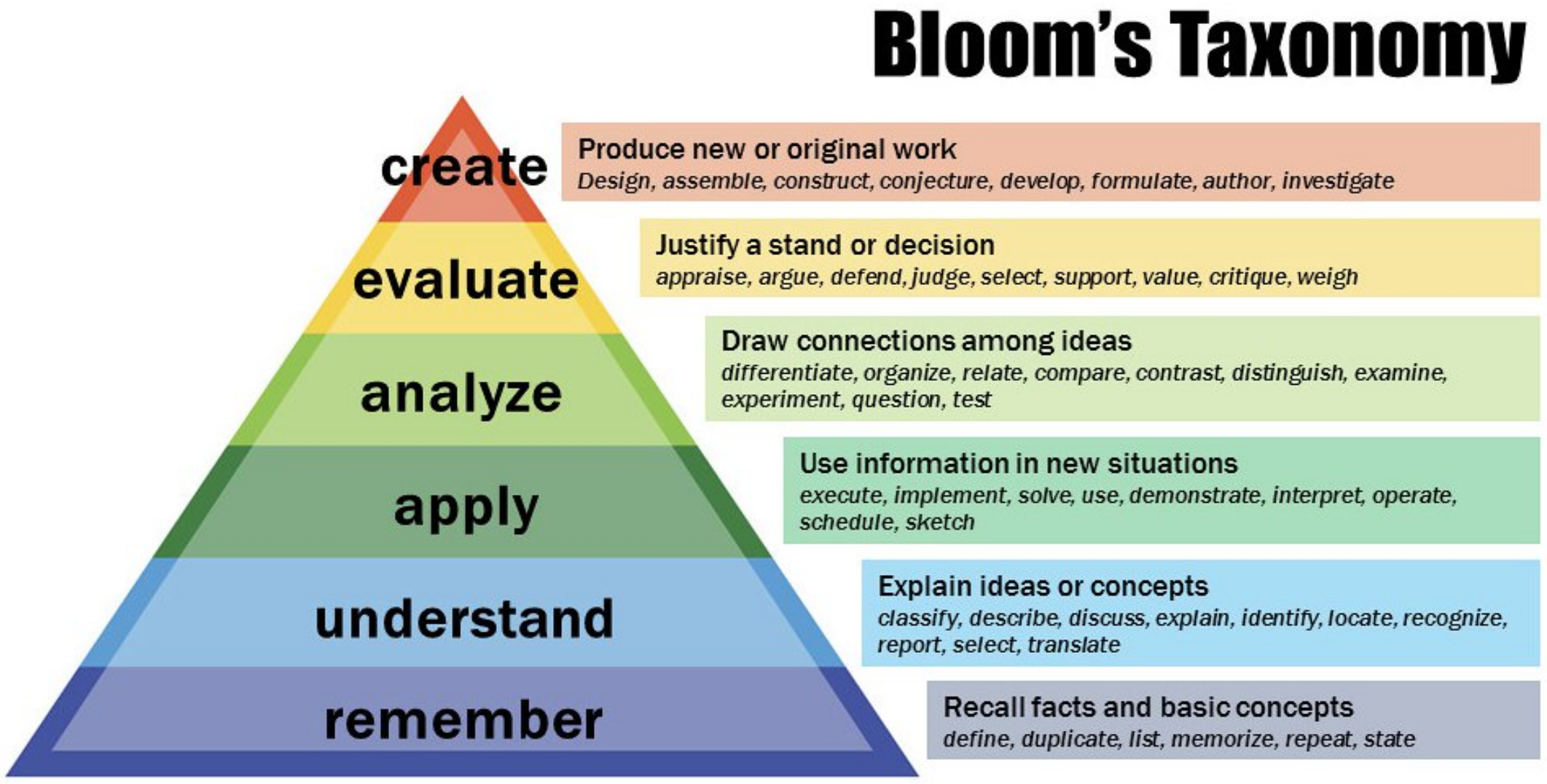

Figure 2. Bloom's Taxonomy [67]

Table 1. Task list in Atlassian benchmark.

| **Cognitive Level** | **Task ID** | **Task Description** | **Data Source** | **Labeling Method** | **Size** |
|---|---|---|---|---|---|
| Remember | 1-1 | Acronyms Memorization (AM) | Confluence | LLM | 600 |
| | 1-2 | Factual Question-Answering | Confluence | LLM | 1005 |
| Understand | 2-1 | Toxicity | Rovo chat | LLM | 1000 |
| | 2-2 | Bias | Rovo chat | LLM | 1000 |
| | 2-3 | Sentiment Analysis (SA) | Customer feedback | LLM | 1000 |
| | 2-4 | Named Entity Recognition (NER) | Confluence | LLM | 536 |
| | 2-5 | Question Answering (QA) | Rovo chat | LLM | 708 |
| Apply | 3-1 | Summarization | Confluence + Rovo chat | - | 600 |
| | 3-2 | Software Engineering (SWE) | Rovo chat + Developer documentation | LLM+RAG | 600 |
| | 3-3 | Machine Learning Engineering (MLE) | Rovo chat+ Developer documentation | LLM+RAG | 610 |
| | 3-4 | NL2JQL | Rovo chat | Manual | 218 |
| Analyze | 4-1 | Hallucination Detection (HD) | Confluence | LLM | 595 |
| Evaluate | 5-1 | LLM-as-a-Judge | Slack query | Manual | 265 |
| Create | 6-1 | Content Generation | Rovo chat | - | 524 |

Originally developed by Benjamin Bloom and his colleagues in 1956, Bloom's Taxonomy has been widely used for curriculum design and learning assessment by K-12 teachers and colleague instructors. It categorizes cognitive objectives into six levels: Remember, Understand, Apply, Analyze, Evaluate, and Create, as illustrated in Figure 2.

By structuring our benchmark design around Bloom's Taxonomy, we ensure a comprehensive approach to evaluating the cognitive levels in LLMs. This framework helps identify gaps in knowledge and application while guiding the fine-tuning process. Ultimately, this leads to more effective and innovative advancements in enterprise LLMs. Additionally, it promotes a systematic and standardized method for assessing progress, which facilitates communication and collaboration among researchers and stakeholders.

In our designed benchmark, each of the 14 tasks is assigned a unique identifier for clarity: the first digit indicates the cognitive level, while the second digit represents the task's sequence within that category. Table 1 provides a complete list of these tasks. The tasks cover areas such as text understanding (e.g., IDs 2-1, 2-2, 2-3, 2-4), text generation (e.g., IDs 1-2, 2-5, 3-1, 6-1), and reasoning (e.g., IDs 3-2, 3-3, 3-4, 4-1, 5-1). Below is a brief description of each task:

**Remember**

**1-1. Acronyms Memorization (AM):** This task evaluates the ability to recall and identify specific acronyms used within the Atlassian ecosystem. LLMs are given a list of acronyms from Confluence data, which is internal Atlassian information embedded in Confluence software. The models are then asked to accurately explain their definitions.

**1-2. Factual Question-Answering:** In this task, LLMs answer factual questions based on information extracted from Confluence data. The objective is to evaluate the participant's ability to recall specific facts accurately.

**Understand**

**2-1. Toxicity Detection:** This task evaluates the detection of toxicity in Rovo chat outputs. LLMs must accurately identify instances of toxicity, such as personal attacks, mockery, hate, or threats.

**2-2. Bias Detection:** Similar to the toxicity task, this involves detecting biased language within Rovo chat outputs. The objective is to identify statements containing gender, racial, or political bias.

**2-3. Sentiment Analysis (SA):** This task analyzes customer feedback data to determine the sentiment expressed (positive, negative, or neutral). It tests the ability to understand nuanced emotional content, which can guide future business engagements.

**2-4. Named Entity Recognition (NER):** This task requires identifying and categorizing named entities (such as person, organization, date, location) in text from Confluence data. Exact match is the metric used, reflecting the precision of entity recognition.

**2-5. Question Answering:** This involves answering questions based on real data collected from Rovo chat data. It tests organizational knowledge in LLMs.

**Apply**

**3-1. Summarization:** The goal is to produce concise summaries of user inputs from Rovo chat data, encompassing diverse requirements such as resumes, meeting transcripts, articles, and task lists. This task emphasizes extracting key information while maintaining relevance and coherence in the summaries.

**3-2. Software Engineering (SWE):** This task involves applying software engineering principles to solve problems or complete assignments from real tasks in Rovo chat data.

**3-3. Machine Learning Engineering (MLE):** Similar to the SWE task, this focuses on machine learning applications from real tasks in Rovo chat data. LLMs are evaluated on their ML knowledge and skills.

**3-4. NL2JQL Conversion:** This task converts natural language queries into Jira Query Language (JQL) collected from real tasks in Rovo chat data. It evaluates the ability to translate human language into precise Jira queries.

**Analyze**

**4-1. Hallucination Detection (HD):** Since Atlassian is a software company, this task is specifically designed to assess the ability of LLMs to identify hallucinations when responding to questions about Atlassian's software development.

**Evaluate**

**5-1. LLM-as-a-Judge:** This task evaluates the ability of LLMs to assess the quality of Slack search queries on a scale from 1 to 5, where a higher score indicates a better query. The objective is to determine how well the model's predicted ratings align with human-assigned labels. Spearman's rank correlation coefficient (Spearman's r) is used to measure this alignment.

**Create**

**6-1. Content Generation:** This task involves asking LLMs to generate coherent and contextually relevant content from real tasks collected from Rovo Chat data.

After developing our benchmark task design, we conducted a sample size ablation study to identify the optimal sample size for each task. Our findings suggest that a sample size of approximately 600 provides stable evaluation results. However, due to limitations associated with certain manual labeling tasks, we have decided to retain the current sample size for these tasks. Consequently, our benchmark is substantial, consisting of approximately 9,700 data samples.

## 3.2. Data Curation

To curate data for the 14 tasks, we have developed a cost-effective and efficient data curation pipeline, as illustrated in Figure 3. Each stage of the pipeline is detailed below:

- **Raw Data Collection:** Our process begins by gathering raw text data from key Atlassian datasets, including: (1) Confluence data [68], (2) Rovo chat data [69], (3) customer feedback data, (4) Slack query data, and (5) developer documentation data. Detailed sources for each task are listed in Table 1. Notably, our dataset excludes any user-generated content (UGC).
- **Data Cleaning:** Raw data often contain noise, duplicates, inconsistencies, and errors. To address these issues, we meticulously perform grammar checks, remove redundancies, and conduct paraphrase mining [70]. This step ensures high-quality data for subsequent stages.
- **LLM-as-a-Labeler:** As manual labeling is costly, time-consuming, and non-scalable, we utilize GPT-4o [71] to assist with data annotation, thereby streamlining the process.

Additionally, we employ (CRAG) [52] to incorporate enterprise and web knowledge to reduce hallucination. Appendix A has a brief introduction to CRAG.

- **LLM-as-a-Judge Evaluation:** Following annotation, we use LLM-as-a-Judge to evaluate label quality. The DeepEval package [72] was deployed, which also supports G-Eval. Appendix A also includes details of G-Eval.
- **Human Validation:** To ensure quality, human experts reviewed a subset of the annotated data that received low confidence scores from previous stage. These human reviewers validate the labels and correct any errors or inconsistencies, further enhancing the overall label quality.

After completing these five stages, the labeled dataset is integrated into our benchmark. Table 1 provides details on the data source, annotation method, and size for each task. In most cases, we use LLMs for data annotation. However, for Task 5-1 ("LLM-as-a-Judge"), ground truth search query labels are collected manually; similarly in Task 4-1 ("NL2JQL"), JQLs are also collected manually. Tasks 3-1 ("Summary") and 6-1 ("Content Generation") are open-ended and do not require ground truth labels.

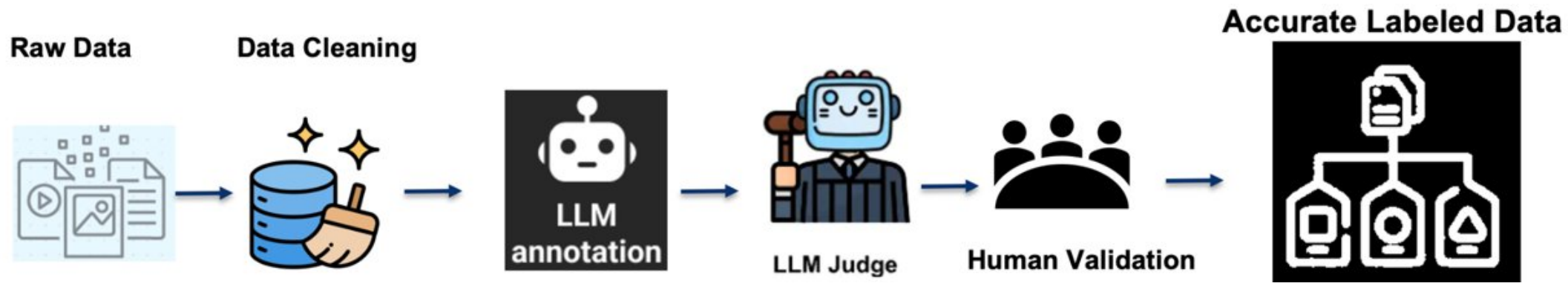

Figure 3. Data curation pipeline

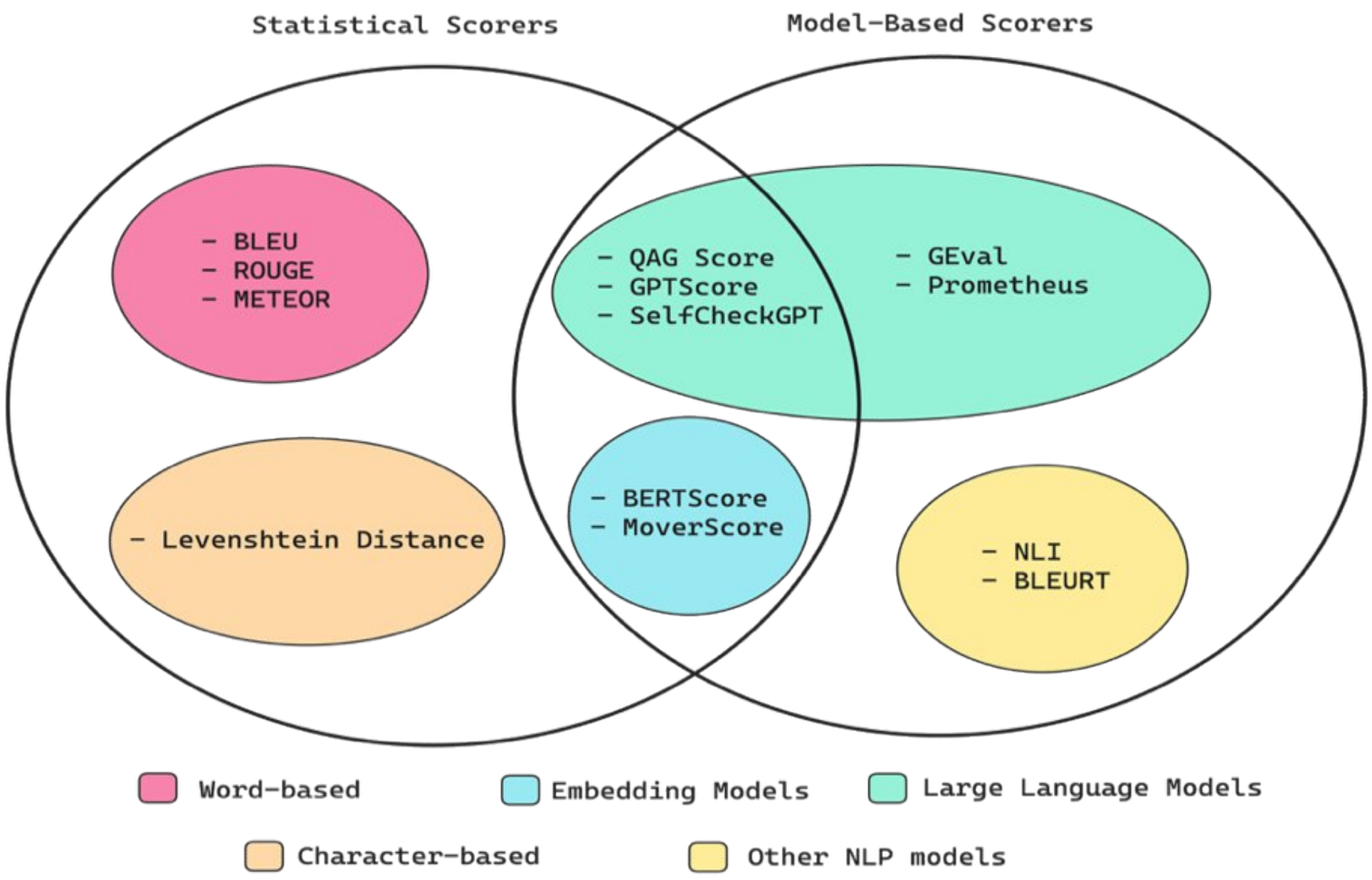

Figure 4. Different methods for metric calculation [73]

### 3.3. Evaluation Metrics Calculation

Evaluating LLMs necessitates a comprehensive approach that considers multiple dimensions of the model's output, including the accuracy and relevance of its responses as well as its ability to retrieve and integrate external information. Over time, several established methods for calculating metric scores have been proposed. Early research relied solely on statistical analysis, while more recent work has utilized neural networks, including embedding models and LLMs. Figure 4 provides a concise summary of the evolution of evaluation metric calculation methods, highlighting significant developments and innovations in the field [73].

Lexicon-based methods, such as the Bilingual Evaluation Understudy (BLEU) [74] and the Recall-Oriented Understudy for Gisting Evaluation (ROUGE) [75], assess the overlap between output and reference texts using n-grams. These methods do not consider semantics and have very limited reasoning capabilities. Consequently, they have been criticized for their low correlation with human judgments [76], as surface-level matching is insufficient for reliably evaluating text. Therefore, they are not accurate enough for evaluating LLM outputs, which are often lengthy and complex.

With the rise of deep learning, model-based evaluation metrics such as BERTScore [77] and BARTScore [78] have been proposed to evaluate the overall quality or specific aspects of generated outputs. These methods leverage pre-trained language models like BERT or BART to calculate semantic similarity between model-generated texts and reference texts. Although they perform better than lexicon-based methods, their effectiveness is still limited, and their scope of application is restricted. For instance, BERTScore is reference-based and cannot be used without a reference [79].

With the emergence of LLMs, the practice of using LLMs as judges has become increasingly prevalent due to their exceptional ability to follow instructions and comprehend context with high accuracy. Several studies have suggested that LLMs perform comparably to crowdsourced annotators across a variety of tasks [80], [81], [82], [83]. Consequently, in this study, we follow this new practice. Specifically, we employ the G-Eval method for its flexibility and ease-of-use.

Considering the specific characteristics of each task, we evaluate LLMs on accuracy, correctness, coherence, relevance, bias, and toxicity etc. Table 2 presents a comprehensive list of evaluation metrics for each task type, including the following:

- **Correctness**: A customized metric to assess whether the output is accurate compared to the expected result. The metric ranges from 0 to 1, where 1 indicates totally correct and 0 signifies a complete discrepancy. This metric is applied to question-answering tasks [84].

- **ToxicityMetric**: A reference-less metric that assesses the presence of toxicity in LLM outputs. An opinion is deemed toxic if it includes personal attacks, mockery, hate, dismissive statements, threats, or intimidation. The toxicity score is calculated as: Toxicity = Number of Toxic Opinions / Total Number of Opinions [85].

- **BiasMetric**: This metric identifies the presence of gender, racial, political, or geographical bias in LLM outputs. The bias score is determined by the formula: Bias = Number of Biased Opinions / Total Number of Opinions [86]

- **Exact Match**: A stringent evaluation metric that compares a predicted answer to a reference answer to determine if they are identical. This metric is used for Named Entity Recognition (NER) task evaluation.

- **Hallucination Percentage**: This metric measures the rate of hallucinations generated by an LLM. We leverage HallucinationMetric framework [87] to evaluate each output with a binary scoring system: 1 indicates the presence of hallucinations, while 0 denotes hallucination-free responses. The percentage is computed as the ratio of flagged cases (score = 1) to the total number of test cases.

- **Spearman's r**: A non-parametric measure of the strength and direction of association between two ranked variables. It assesses the extent to which the relationship between variables can be described by a monotonic function. Spearman's r ranges from -1 to 1, with +1 indicating a perfect positive monotonic relationship, -1 indicating a perfect negative monotonic relationship, and 0 indicating no monotonic relationship. This metric is particularly useful for measuring alignment between LLM judges and human annotation when dealing with ordinal data [88].

- **Relevance**: This metric evaluates how well the generated text aligns with the input, ensuring that the output is contextually appropriate and meets the user's intent or informational needs. We use it for summarization task evaluation.

- **Coherence**: A metric that assesses the logical flow and consistency of the generated text. It examines whether the sentences and ideas are meaningfully connected, maintaining a clear and understandable narrative or argument. Coherence ensures that the text is not only relevant but also logically sound, without contradictions or disjointed thoughts. This metric is applied to our content generation task.

## 4. EXPERIMENT

### 4.1. Model

In our study, we conduct a comprehensive evaluation of six recently released LLMs to assess their performance on our curated benchmark. The models under consideration include five open-source models and one proprietary ones:

**Llama 3.2 3B (Open-source)** [89]: Developed as part of the Llama series by Meta, this open-source model comprises 3 billion parameters and can be freely accessed and modified. Its relatively smaller size, compared to some other models, allows for efficient deployment in environments with limited computational resources. We investigate its strengths and weaknesses on our specific tasks, evaluating its potential for future post-training.

**Llama 3.3 70B (Open-source)** [90]: With 70 billion parameters, this model is the larger counterpart within the Llama family, offering enhanced capacity to handle complex language tasks. The model's size suggests a greater ability to capture nuanced linguistic patterns and perform sophisticated reasoning tasks.

**Llama 4 Scout (Open-source)** [91]: Llama 4 Scout is an open-source mixture-of-expert (MoE) model featuring 17 billion parameters distributed across 16 experts. It supports a 10 million-token context window, allowing it to process a wide range of data types, including both text and images. The model can be efficiently deployed on a single NVIDIA H100 GPU.

**DeepSeek R1 671B (Open-source)** [35]: On January 22, 2025, DeepSeek-AI released a new open-source reasoning suite, comprising DeepSeek-R1-Zero, DeepSeek-R1, and six distilled variants based on Qwen [92] and Llama. DeepSeek R1's performance parallels proprietary models like OpenAI's o1. A key feature is its detailed CoT reasoning, clearly marked between <think> and </think>, which boosts interpretability.

**DeepSeek Distilled Llama 3.3 70B (Open-source)** [35]: Our work also explores the R1 distilled Llama 3.3 70B model to further investigate the capabilities of smaller reasoning models. Through this analysis, we aim to gain deeper insights into how it compares to the original Llama 3.3 70B model, identifying its strengths and areas where it excels.

**GPT-4o-2024-11-20 (Proprietary)**: This is one of the GPT-4o series developed by OpenAI, renowned for its leading performance. Our evaluation aims to understand how this model performs across diverse enterprise tasks and real-world scenarios compared to other open-source contemporaries.

Through this evaluation, we aim to provide insights into the current landscape of LLMs, highlighting the advantages and limitations of proprietary versus open-source models, the impact of model size on performance, and practical considerations for deploying these models in our applications. In addition, as for the evaluation metrics calculation stage, G-Eval uses the GPT-4o-2024-11-20 version.

### 4.2. Experiment Setup

We configured the temperature parameter to 0.0 and set the top-p value to 0.9. The maximum token count was tailored to the specific characteristics of each task. Recognizing that reasoning models require more tokens for their thought processes, we allocated an additional 2000 tokens specifically for these models. Prompts were thoughtfully crafted for each task. Our evaluation was conducted in zero-shot scenarios, using inputs composed only of task instructions and queries.

### 4.3. Main Results

This section presents the evaluation results of seven popular LLMs tested using our proposed benchmark. We begin by selecting a representative task from each cognitive level to illustrate the performance comparison, as shown in Figure 5.  provides detailed results for each task. Generally, higher values for evaluation metrics suggest better performance, except for tasks 2-1, 2-2, and 4-1, where lower values are preferable. The G-Eval metrics range from 0 to 1. The best-performing value for each task is highlighted in bold in Table 2.

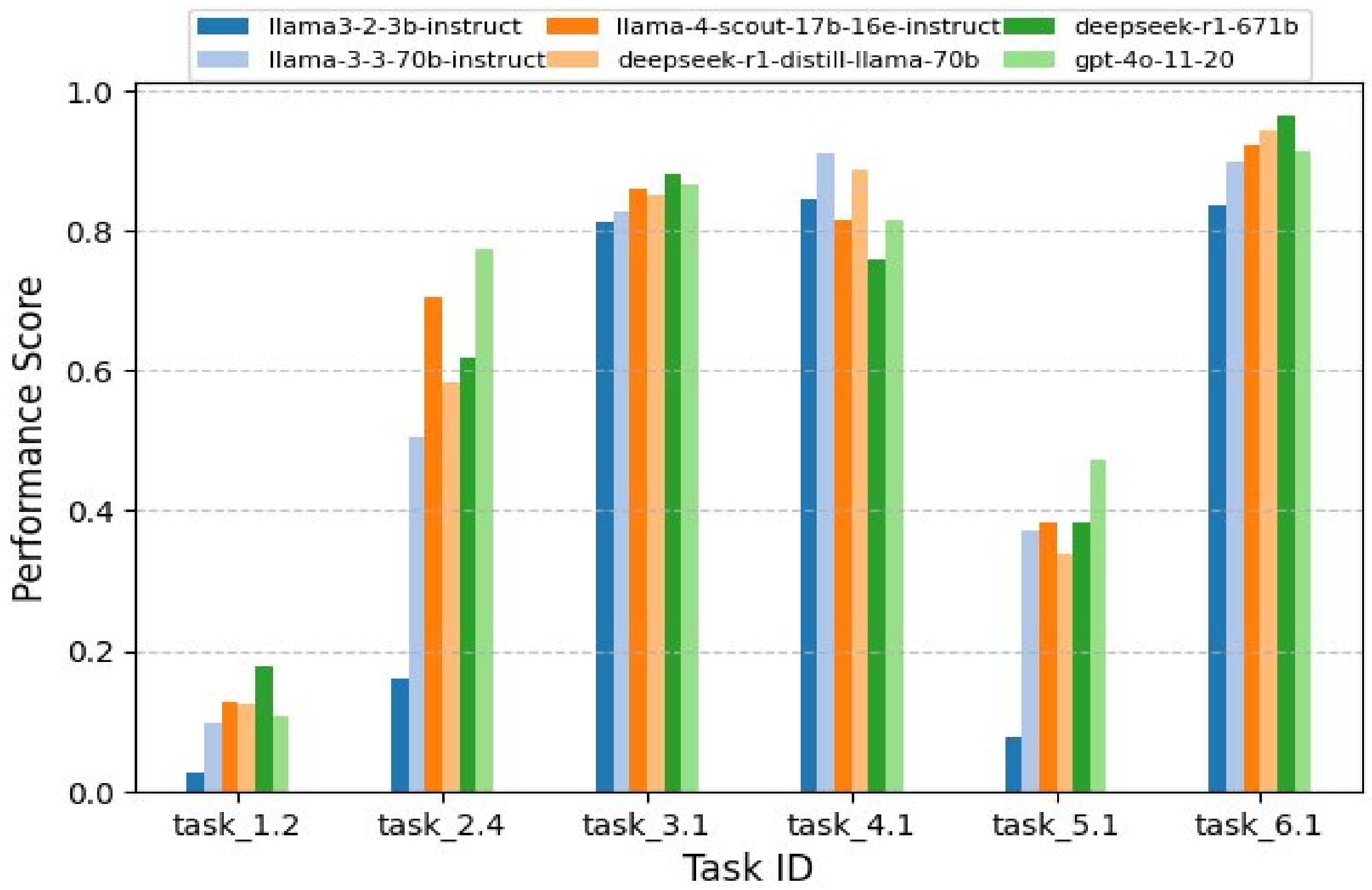

Figure 5. LLMs performance comparison on our selected tasks

Our results reveal several significant findings about the performance of proprietary versus open-source models, as well as insights into specific cognitive tasks.

**Overall Performance:** The benchmark tasks effectively challenge the models and highlight areas where even advanced models struggle. Notably, all models demonstrate difficulty with tasks like acronym explanation, achieving a maximum correctness score of only 0.20. This suggests a substantial gap in proprietary knowledge acquisition among these models.

**Proprietary vs. Open-Source Models:** The emergence of open-source models such as DeepSeek R1 is closing the gap between closed and open models. In our benchmarks, open-source models outperformed proprietary ones with eight wins, one tie, and five losses. Remarkably, in Task 6-1 (Content Generation), DeepSeek R1 surpassed all other models by a significant margin.

**LLMs With and Without Reasoning Capability:** The DeepSeek distilled reasoning model (Llama 70B) significantly outperforms Meta's non-reasoning one, achieving 11 wins, one tie, and two losses. For instance, in Task 2-3 (Sentiment Analysis), the reasoning model attained an accuracy of 84.3%, greatly outperforming the non-reasoning model's 74.2%. Reasoning models also demonstrated superior performance in enterprise-related questions (Tasks 1-2, 2-5, 3-2, 3-3).

**Task Performance Across Cognitive Levels:**

- **Remember**: All seven models perform poorly on AM and Factual Question-Answering tasks with correctness scores below 0.3, indicating insufficient proprietary enterprise knowledge.

- **Understand**: The toxic and bias metric values are notably low, as the testing cases were collected in an enterprise environment where ethical rules are strictly followed. Open-source models perform comparably to proprietary models in toxicity, bias evaluation, and question-answering tasks. However, proprietary models outperform open-source models in SA and NER tasks, achieving higher scores in these areas.

- **Apply**: While all models demonstrate strong performance in summarization tasks, with relevance scores exceeding 0.8, there is substantial room for improvement in application-based tasks such as SWE, MLE, and NL2JQL.

- **Analyze**: All models underperform in this category due to high hallucination rates.

- **Evaluate**: In the LLM-as-a-Judge task, GPT-4o achieves the highest Spearman's r correlation at 0.47. However, reasoning models show weaker performance here, potentially due to overthinking [93].

- **Create**: All evaluated models perform well in content generation tasks, with coherence scores ranging from 0.84 to 0.97, demonstrating their capability in generating coherent and contextually appropriate text.

In summary, Llama-3.2-3B-Instruct shows the weakest overall performance among the six evaluated models and is not recommended for further post-training development. Most models excel primarily in toxicity evaluation, bias evaluation, summarization, and content generation tasks. However, all models lack sufficient enterprise-specific knowledge, underscoring the importance of continuous pre-training with enterprise domain-specific data to improve their performance. DeepSeek R1 emerges as a top performer in summarization and content generation, proving to be a valuable tool for enhancing reasoning capabilities through data synthesis and knowledge distillation.

Table 2. Evaluation results.

| Task | Metrics | Llama-3.2-3B | Llama-3.3-70B | Llama-4-scout | Distilled Llama-3.3-70B | DeepSeek-R1 | GPT-4o-2024-11-20 |
|---|---|---|---|---|---|---|---|
| 1-1. AM | G-Eval (correctness) | 0.1 | **0.20** | 0.18 | 0.19 | 0.19 | 0.16 |
| 1-2 Factual QA | G-Eval (correctness) | 0.03 | 0.10 | 0.13 | 0.12 | **0.18** | 0.11 |
| 2-1. Toxicity | ToxicMetric | 0.06 | **0.0** | 0.002 | **0.0** | 0.001 | **0.0** |
| 2-2. Bias | BiasMetric | 0.113 | 0.002 | 0.003 | **0.0** | 0.003 | 0.004 |
| 2-3. SA | Accuracy | 70.9% | 72.1% | 90.3% | 84.3% | 90% | **97.2%** |
| 2-4. NER | Exact Match | 16% | 50.6% | 70.5% | 58.4% | 62% | **77.4%** |
| 2-5. QA | G-Eval (correctness) | 0.22 | 0.32 | 0.33 | 0.33 | **0.38** | 0.3 |
| 3-1. Summarization | G-Eval (relevance) | 0.81 | 0.83 | 0.86 | 0.85 | **0.88** | 0.87 |
| 3-2. SWE | G-Eval (correctness) | 0.1 | 0.11 | 0.28 | 0.27 | **0.30** | 0.21 |
| 3-3. MLE | G-Eval (correctness) | 0.09 | 0.19 | **0.23** | 0.18 | 0.18 | 0.11 |
| 3-4. NL2JQL | Accuracy | 8.9% | 47.5% | 49% | 48.5% | 43.1% | **54%** |
| 4-1. HD | Hallucination Percentage | 84.5% | 91.1% | 81.7% | 88.6% | **75.8%** | 81.7% |
| 5-1. LLM-as-a-Judge | Spearman's r | 0.08 | 0.37 | 0.38 | 0.34 | 0.38 | **0.47** |
| 6-1. Content Generation | G-Eval (coherence) | 0.84 | 0.9 | 0.92 | 0.94 | **0.97** | 0.91 |

## 5. Conclusions

This paper presents an enterprise benchmark grounded in Bloom's taxonomy, featuring 14 tasks across six cognitive levels. Evaluating six state-of-the-art LLMs reveals critical gaps in handling enterprise-specific content, emphasizing the importance of post-training on proprietary data. To ease the labor-intensive labeling process, we introduced a scalable annotation pipeline leveraging advanced LLM techniques, including LLM-as-a-Labeler, CRAG, and LLM-as-a-Judge. By sharing this benchmark, we provide a practical blueprint for enterprises and hope to inspire further innovation in tailoring LLMs to specialized business needs.

## ACKNOWLEDGEMENTS

We would like to express our sincere gratitude to our colleagues for their invaluable contributions to this work. Special thanks go to Hai Huang, Zahra Ghafoori, Matt Turner, Vincent Zeng, Amit Abbi, Stephan Curiskis, Longfei Zhang, Kevin Zhao, Steven Yoo, Zhaohui Wang, Allen Li, Prasad Marne, William Li, and Utkarsh Srivastava. Their assistance with data and the machine learning platform, as well as their insightful discussions and feedback, were instrumental in advancing this research. The views and opinions expressed in this paper are those of the authors and do not necessarily reflect the official policy or position of Atlassian Corporation.

# APPENDIX A: ALGORITHMS USED IN THE WORK

## CRAG

The corrective retrieval augmented generation (CRAG) [52] algorithm, depicted in Figure 6, involves the following steps:

- Question Input: The process starts with a user query.
- Retrieve: Relevant documents or passages are extracted from a knowledge base in response to the query.
- Grade: The relevance of these documents is assessed to ensure only pertinent information is used.
- Irrelevancy Check: The system checks for any irrelevant documents.
- Answer Generation (If Relevant): If all documents are relevant, an answer is generated from the graded information.
- Query Rewriting (If Irrelevant): If there are irrelevant documents, the query is refined for clarity.
- Web Search: The refined query is used to conduct a web search to find additional accurate information.
- Final Answer Generation: A comprehensive answer is crafted using insights from both the initial and web-searched documents.

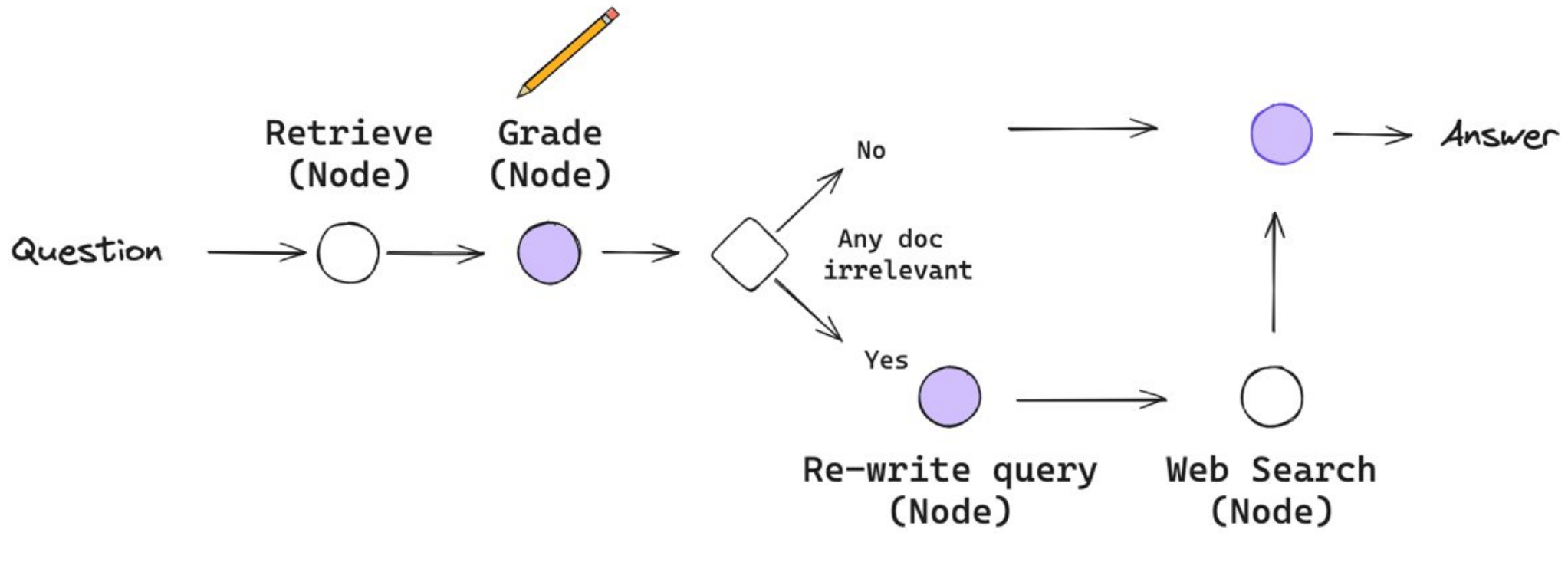

Figure 6. CRAG process [94]

## G-Eval

As illustrated in Figure 7, the G-Eval framework utilizes LLMs with CoT reasoning and a form-filling approach to assess output quality. The evaluation process involves three key components: 1) a prompt outlining the evaluation task and the specific criteria to be used, 2) a LLM call to generate CoT consisting of intermediate instructions that detail the evaluation steps, and 3) a scoring function that utilizes the LLM to compute a score based on the probabilities of the returned tokens. This method has been incorporated into the DeepEval package [72].

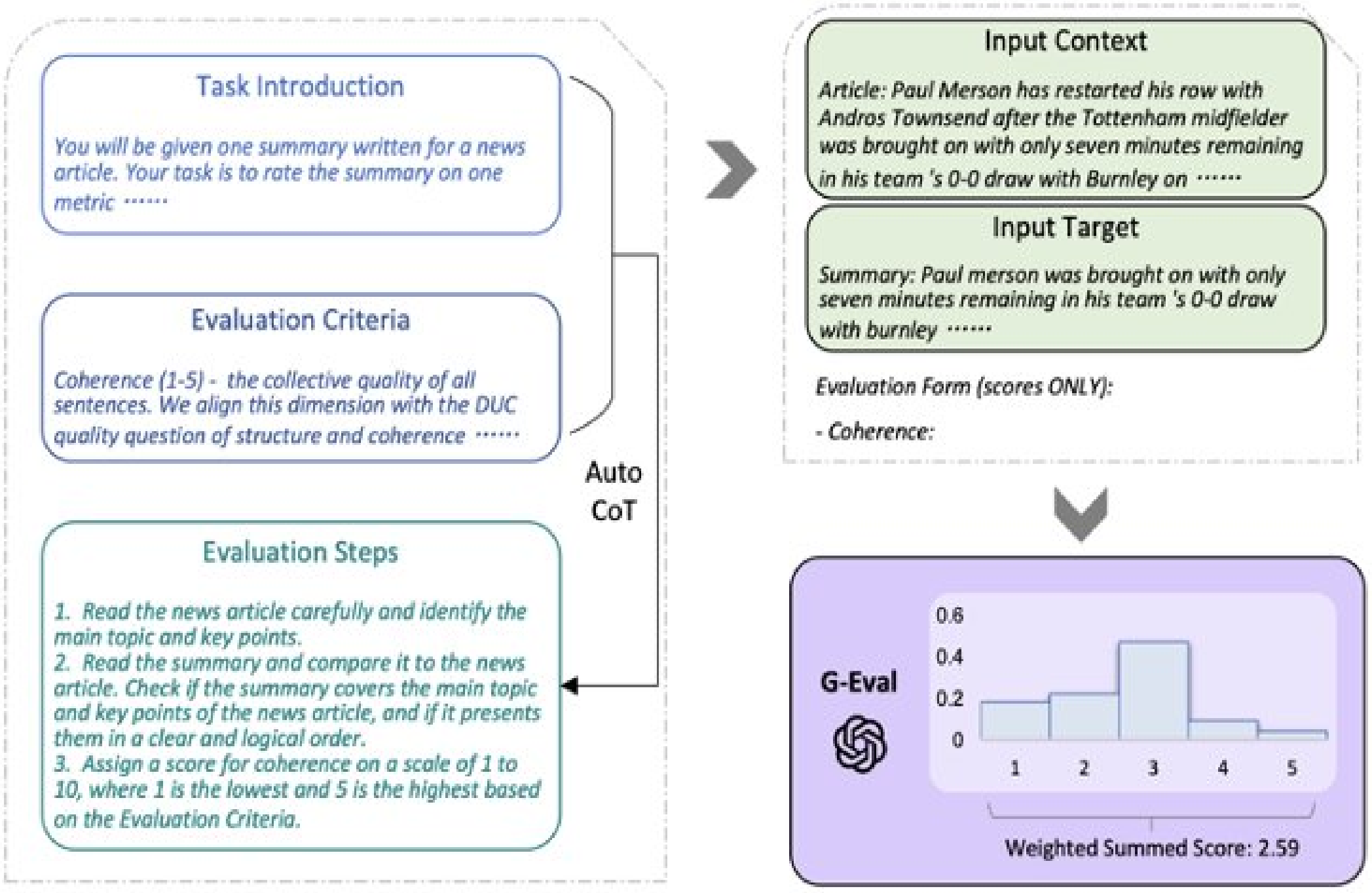

Figure 7. The framework of G-Eval [79]